%% file: main.tex
\documentclass[letterpaper]{article}
\PassOptionsToPackage{table}{xcolor}

\usepackage[preprint]{aaai2027}

\usepackage{natbib}
\usepackage[hyphens]{url}
\usepackage{graphicx}
\usepackage{amsmath}
\usepackage{amssymb}
\usepackage{booktabs}
\usepackage{multirow}
\usepackage{xspace}
\usepackage{enumitem}
\usepackage{array}
\usepackage{xcolor}
\usepackage{listings}

\newcommand{\method}{\mbox{GapSight}\xspace}

\newcommand{\lossgap}{loss-gap\xspace}
\newcommand{\sixavg}{Six-Bench Avg\xspace}

\definecolor{TableHeader}{HTML}{F6F8FA}
\definecolor{TableOurs}{HTML}{E4F0FF}
\definecolor{HelaPromptBlue}{HTML}{EAF8FE}
\definecolor{HelaPromptPink}{HTML}{FCF3FB}

\lstdefinestyle{helapromptbase}{
    basicstyle=\ttfamily\footnotesize,
    frame=single,
    framerule=0.45pt,
    rulecolor=\color{black},
    framesep=6pt,
    framexleftmargin=1.8em,
    xleftmargin=0pt,
    xrightmargin=0pt,
    aboveskip=0.45em,
    belowskip=1.0em,
    numbers=left,
    numberstyle=\tiny,
    numbersep=8pt,
    breaklines=true,
    breakatwhitespace=true,
    breakautoindent=false,
    breakindent=0pt,
    columns=fullflexible,
    keepspaces=true,
    showstringspaces=false
}
\lstdefinestyle{promptblue}{
    style=helapromptbase,
    backgroundcolor=\color{HelaPromptBlue}
}
\lstdefinestyle{promptpink}{
    style=helapromptbase,
    backgroundcolor=\color{HelaPromptPink}
}

\title{Learning to Look Again: Loss-Gap Supervision for Free-form Crop Routing in Vision-Language Models}

\author{
    Jinchang Zhu$^{1,a}$,
    Rong Fu$^{2}$,
    Yi Ding$^{1}$,
    Chenghao Wu$^{1}$,
    Ying Liu$^{1}$,
    Menglin Yang$^{1,b}$\thanks{Corresponding author.}
}
\affiliations{
    $^{1}$The Hong Kong University of Science and Technology (Guangzhou)\\
    $^{2}$University of Macau\\
    \texttt{$^{a}$jzhu997@connect.hkust-gz.edu.cn \quad $^{b}$menglinyang@hkust-gz.edu.cn}
}

\begin{document}
\maketitle

\begin{abstract}
Vision-language models (VLMs) fail many detail-centric questions for a concrete reason: the answer is visible in the image, yet lost after the image is compressed into a low-resolution global view. Allocating more visual tokens to every query improves some OCR and document cases, but it spends computation indiscriminately and can disturb tasks that rely on global context. We propose \method, a framework for learning \emph{visual re-reading}: a VLM first takes a global glance, then selectively returns to a free-form region when the question calls for local evidence. The supervision comes from the target model's own failure signal. Offline, we compare answer loss or multiple-choice option margin under a global-only view and candidate crop-augmented views; crops that improve the target answer become model-specific review labels. A lightweight free-form crop router distills these labels into a one-shot inference policy that predicts whether to review, expected utility, and a continuous crop box from the global state. Across LLaVA-1.5-7B, InternVL2.5-8B, and Qwen2-VL-2B-Instruct, \method improves the Base no-zoom baseline on six benchmarks spanning OCR, documents, charts, infographics, VStarBench, and MME-RealWorld-Lite. On InternVL2.5-8B, \method raises the six-benchmark average from 52.25 to 64.29, above CropVLM (57.16), ViCrop (55.84), and ZoomRefine (54.43). Mechanism analyses show that the router rescues concrete wrong answers, adapts its action rate by task, and forms a favorable token-performance profile. These results position loss-gap supervision as a practical route to teaching VLMs when and where to look again.
\end{abstract}

\input{sections/01_introduction}
\input{sections/02_related_work}

\input{sections/03_method}
\input{sections/04_experiments}
\input{sections/05_analysis}
\input{sections/07_conclusion}

\clearpage
\bibliography{references}

\end{document}

%% file: sections/01_introduction.tex
\section{Introduction}

Vision-language models (VLMs) are increasingly used as general visual assistants for documents, charts, infographics, screenshots, and real-world scenes. A central bottleneck remains unresolved: the visual evidence required by a question is often more precise than the evidence preserved by the model's input representation. A global image view gives the model layout and context, yet it can erase the local signal that determines the answer. Receipt text blurs, chart values collapse, infographic labels shrink, and small objects become ambiguous. In such cases the answer is present in the image but inaccessible after visual compression.

The natural engineering response is to spend more visual tokens. High-resolution tiling, repeated global views, and always-on crop augmentation expose more pixels and can substantially improve OCR-heavy tasks. They also allocate local evidence indiscriminately. Many questions are already solved from the global view, and some benchmarks reward global spatial context more than magnified local texture. A stronger VLM inference procedure supports \emph{visual re-reading}: first form a global interpretation, then return to a precise region when the question requires evidence that the global view failed to preserve.

The key challenge is supervision. Useful crops are model-specific. The same region can help one VLM and fail to help another because the benefit depends on the visual encoder, image tokenizer, language decoder, instruction tuning, prompt format, and answer parser. Human boxes capture semantic objects or text spans, but the VLM may need a wider region containing layout, units, legends, or neighboring fields. Training-free crop methods avoid annotation but pay for test-time search or rely on external signals that only indirectly reflect answer quality.

We propose \method, a framework for learning visual re-reading from loss-gap supervision. The core idea is simple: the target VLM can reveal useful visual evidence through its own answer behavior. During offline label mining, we compare the target answer under a low-resolution global view with the same answer under candidate crop-augmented views. For generated answers, the signal is the reduction in answer-span negative log-likelihood. For multiple-choice benchmarks, the signal is the increase in correct-option margin. A candidate crop that produces a large positive gap is useful evidence for this VLM on this example. A candidate crop that fails to improve the answer is weak evidence. These measured gaps become supervision for both \emph{when} to re-read and \emph{where} to re-read.

\begin{figure*}[t]
    \centering
    \includegraphics[width=.98\textwidth]{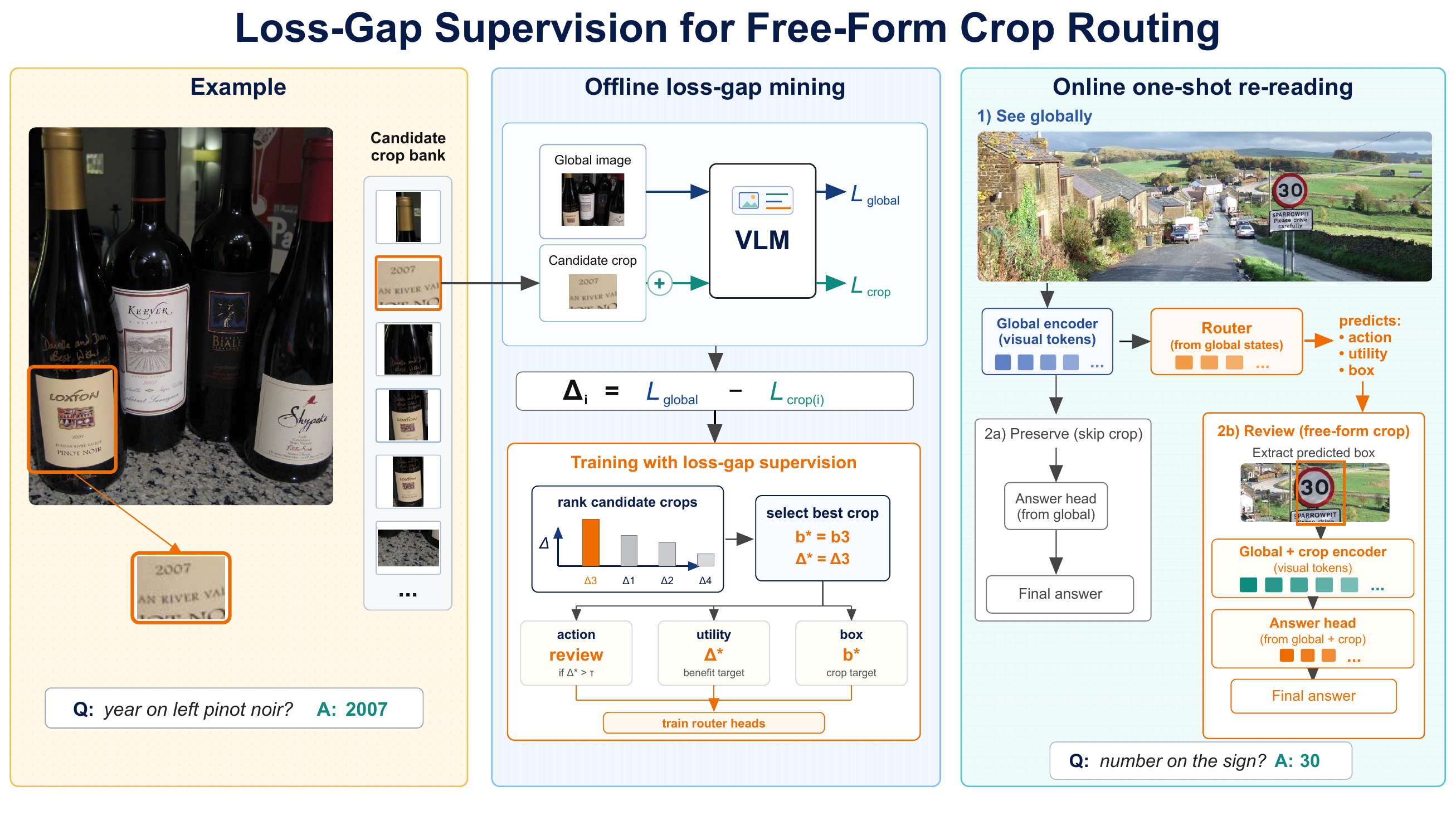}
    \caption{Overview of \method. Offline loss-gap mining probes the target VLM with a global view and candidate crop-augmented views, converting answer-loss or option-margin improvements into supervision for action, utility, and free-form box prediction. At inference time, the free-form crop router predicts from the global state whether to preserve the preview or inject one crop for visual re-reading.}
    \label{fig:overview}
\end{figure*}

\method trains a lightweight free-form crop router attached to the VLM. From hidden states computed on the global view, the router predicts whether to preserve or review, a scalar utility, and a continuous bounding box. At inference time, the model first processes the global preview. If the router chooses review, one crop is injected and the final answer is generated from the global-plus-crop context. The crop decision is made before crop tokens are available, so the router learns a genuine allocation policy from the global state. The free-form box lets the selected region follow text blocks, document fields, chart areas, and object extents.

This framing has three practical consequences. First, supervision is aligned to the target backbone. LLaVA, Qwen, and InternVL produce different gains from the same crop candidates, and \method mines labels separately for each model family. Second, the method cleanly separates \emph{learning to re-read} from \emph{answer generation}. The VLM keeps its ordinary generative interface while the router controls visual evidence allocation. Third, the approach exposes a continuous cost-performance tradeoff: gate and utility thresholds adjust how often the model reviews an image, giving practitioners a controllable visual budget.

We evaluate \method across LLaVA-1.5-7B, InternVL2.5-8B, and Qwen2-VL-2B-Instruct. The evaluation spans TextVQA, DocVQA, ChartQA, InfographicVQA, VStarBench, and MME-RealWorld-Lite, covering OCR, document layout, charts, infographic reading, small-object reasoning, and high-resolution real-world multiple-choice understanding. \method improves the corresponding Base no-zoom settings across all reported model rows. On InternVL2.5-8B, \method raises \sixavg from 52.25 to 64.29, above CropVLM (57.16), ViCrop (55.84), and ZoomRefine (54.43). On LLaVA-1.5-7B, it improves over Base no-zoom by 9.81 points. On Qwen2-VL-2B-Instruct, it raises \sixavg from 44.65 to 56.02.

The aggregate gains come from a learned visual re-reading policy rather than from uniformly adding crop tokens. On InternVL2.5-8B, a transition analysis shows 107 error repairs against 27 regressions across the four VQA benchmarks. The learned gate reviews frequently on text- and infographic-heavy tasks while abstaining more often when global scene context matters. Under this policy, \method reaches 64.29 \sixavg with 391.0 average visual tokens, below the token cost of the external crop baselines.

The paper makes the following contributions:
\begin{itemize}[leftmargin=1.2em,itemsep=0.2em]
    \item \textbf{Loss-gap supervision for visual re-reading.} We introduce a supervision signal that measures whether a candidate crop improves the target VLM's own answer behavior, using answer-NLL reduction for generated answers and option-margin improvement for multiple-choice benchmarks.
    \item \textbf{A one-shot free-form crop router.} We train a lightweight router that predicts, from the global image state alone, whether to review, how useful review is expected to be, and which continuous region to inject before final answering.
    \item \textbf{Cross-backbone evaluation and behavioral evidence.} We instantiate the framework on LLaVA-1.5-7B, InternVL2.5-8B, and Qwen2-VL-2B-Instruct across six benchmarks, showing gains over Base no-zoom and external crop/zoom baselines, task-adaptive gate behavior, and model-specific crop utility.
\end{itemize}

%% file: sections/02_related_work.tex
\section{Related Work}

\paragraph{High-resolution evidence in vision-language models.}
Modern VLMs build on large-scale image-text pretraining and multimodal instruction tuning. CLIP established contrastive visual-language representation learning at web scale~\cite{radford2021clip}; Flamingo, BLIP-2, and PaLI showed that frozen or jointly scaled visual encoders can be coupled with large language models for broad multimodal transfer~\cite{alayrac2022flamingo,li2023blip2,chen2022pali}. Recent open VLM families further improve instruction following, OCR, localization, and high-resolution perception, including LLaVA~\cite{liu2023llava}, Qwen-VL and Qwen2-VL~\cite{bai2023qwenvl,wang2024qwen2vl}, and InternVL2.5~\cite{chen2024internvl}. At the same time, document, chart, and text-rich systems such as Donut, Pix2Struct, MatCha, and mPLUG-DocOwl make clear that many answers depend on fine-grained evidence preserved only at sufficiently high visual resolution~\cite{kim2022donut,lee2023pix2struct,liu2023matcha,ye2023mplugdocowl}. The common bottleneck is visual evidence allocation: a compact global view preserves layout and context, while local detail requires additional tokens or a second look.

\paragraph{Dynamic resolution and visual-token allocation.}
A large body of work improves visual efficiency by changing how pixels become tokens. NaViT packs images with variable aspect ratios and resolutions into a flexible transformer input~\cite{dehghani2023navit}; LLaVA-UHD and Monkey explore high-resolution perception for multimodal LLMs through image partitioning, aspect-ratio handling, and resolution-aware training~\cite{xu2024llavauhd,li2024monkey}. Orthogonal efficiency methods prune, merge, or learn compact token sets: TokenLearner compresses visual inputs into a small number of learned tokens~\cite{ryoo2021tokenlearner}, DynamicViT drops less informative image tokens~\cite{rao2021dynamicvit}, ToMe merges redundant visual tokens~\cite{bolya2023tome}, and recent VLM-specific methods such as FastV and PruMerge exploit the redundancy of visual tokens inside multimodal decoders~\cite{chen2024fastv,shang2024prumerge}. These methods primarily act on image structure, token salience, or model-internal token redundancy. \method addresses a different allocation decision: whether a particular question for a particular VLM benefits from injecting a free-form crop, and where that crop should be. The routing signal is measured by answer consequence rather than by image geometry alone.

\paragraph{Visual search, cropping, and zooming.}
Explicit reinspection has become an important inference primitive for VLMs. V* formulates guided visual search as a mechanism for resolving visual challenges in multimodal LLMs~\cite{wu2024vstar}. ViCrop improves zero-shot VQA by cropping visually relevant regions before answering~\cite{yang2023vicrop}. ZoomEye uses tree-based image exploration to mimic human-like zooming~\cite{shen2024zoomeye}. Zoom-Refine prompts a VLM to localize, zoom, and refine its answer~\cite{yu2025zoomrefine}. CropVLM trains a dedicated cropping model for fine-grained vision-language perception~\cite{carvalho2025cropvlm}. These methods demonstrate that a second visual read can be more valuable than uniformly increasing the full-image resolution. They also expose the central difficulty: a crop policy must know when local magnification helps and when global context should dominate. \method makes this decision trainable from target-model evidence. Offline candidate probing records which crops improve the target VLM's answer loss or correct-option margin, and the resulting labels are distilled into a one-shot router that predicts action, utility, and a continuous box from the global state.

\paragraph{Learning supervision from model behavior.}
Learning from model behavior is a recurring theme in language modeling~\cite{xue2026supervised,xue2026reason}. Human preference comparisons provide reward signals for alignment~\cite{christiano2017preferences,stiennon2020learning,ouyang2022instructgpt}; AI feedback and direct preference optimization make preference learning practical without conventional supervised targets for every decision~\cite{bai2022constitutional,rafailov2023dpo}. Self-improvement methods such as STaR, Self-Refine, and Reflexion use model-generated reasoning, critique, or feedback to turn behavior traces into new training signals~\cite{zelikman2022star,madaan2023selfrefine,shinn2023reflexion}. \method brings this comparison-based view to visual action learning. The measured difference between global-only and crop-augmented answer behavior becomes supervision for a visual decision: preserve the global view or review a region. For generated answers, the signal is answer-span likelihood improvement; for multiple-choice benchmarks, it is correct-option margin improvement. The same loss gap supplies an action label, a utility target, and evidence boxes for free-form crop prediction.

\paragraph{Text-rich and high-resolution evaluation.}
Recent VLM benchmarks increasingly test whether limited visual tokens preserve text, layout, charts, small objects, and real-world high-resolution evidence~\cite{singh2019textvqa,mathew2021docvqa,masry2022chartqa,mathew2022infographicvqa,wu2024vstar,zhang2025mmerealworld}. Broader suites such as OCRBench, MMBench, MMMU, and MathVista reinforce the same pressure toward detail-sensitive evaluation under visual-token budgets~\cite{liu2024ocrbench,liu2023mmbench,yue2024mmmu,lu2024mathvista}. These settings make visual re-reading a concrete allocation problem: a useful method should improve detail-heavy cases, preserve global reasoning, and expose its token cost.

\input{tables/full_benchmark_results}

%% file: tables/full_benchmark_results.tex
\begin{table*}[t]
\centering
\scriptsize
\setlength{\tabcolsep}{1.8pt}
\renewcommand{\arraystretch}{0.94}
\resizebox{0.96\textwidth}{!}{
\begin{tabular}{llcccccccc}
\toprule
\rowcolor{TableHeader}
Backbone & Method & TextVQA & DocVQA & ChartQA & InfoVQA & VStar & MME-Lite & Six Avg & Avg. tokens \\
\midrule
\multirow{5}{*}{LLaVA-1.5-7B}
& Base no-zoom & $58.63$ & $20.87$ & $21.43$ & $17.72$ & 45.18 & 32.71 & $32.76{\pm}0.10$ & 576.0 \\
& ZoomRefine & $62.18$ & $28.73$ & $24.31$ & $24.58$ & 46.47 & 28.39 & $35.78{\pm}0.24$ & 1151.2 \\
& ViCrop & $59.04$ & $27.18$ & $21.76$ & $22.94$ & 51.96 & \textbf{33.88} & $36.13{\pm}0.22$ & 1152.0 \\
& CropVLM & $58.12$ & $27.66$ & $25.37$ & $20.83$ & \textbf{52.64} & 33.66 & $36.38{\pm}0.25$ & 1151.3 \\
\rowcolor{TableOurs}
& \method & $\textbf{71.84}$ & $\textbf{40.28}$ & $\textbf{33.16}$ & $\textbf{27.36}$ & 48.93 & 33.82 & $\textbf{42.57}{\pm}0.33$ & 912.8 \\
\midrule
\multirow{5}{*}{InternVL2.5-8B}
& Base no-zoom & $73.74$ & $55.84$ & $63.61$ & $39.42$ & 51.24 & 29.66 & $52.25{\pm}0.10$ & 256.0 \\
& ZoomRefine & $74.31$ & $60.66$ & $65.74$ & $43.21$ & 53.11 & 29.53 & $54.43{\pm}0.31$ & 462.4 \\
& ViCrop & $80.16$ & $57.91$ & $64.53$ & $44.38$ & 55.42 & 32.61 & $55.84{\pm}0.21$ & 512.0 \\
& CropVLM & $81.73$ & $66.52$ & $63.86$ & $42.71$ & 55.57 & 32.54 & $57.16{\pm}0.25$ & 511.9 \\
\rowcolor{TableOurs}
& \method & $\textbf{87.26}$ & $\textbf{78.11}$ & $\textbf{67.63}$ & $\textbf{53.27}$ & \textbf{58.72} & \textbf{40.76} & $\textbf{64.29}{\pm}0.24$ & 391.0 \\
\midrule
\multirow{5}{*}{Qwen2-VL-2B-Instruct}
& Base no-zoom & $67.08$ & $45.96$ & $50.83$ & $33.41$ & 43.58 & 27.02 & $44.65{\pm}0.10$ & 144.0 \\
& ZoomRefine & $73.62$ & $54.78$ & $47.36$ & $33.94$ & 44.91 & 26.19 & $46.80{\pm}0.27$ & 378.9 \\
& ViCrop & $72.34$ & $55.26$ & $47.11$ & $35.64$ & 47.73 & 26.04 & $47.35{\pm}0.24$ & 400.0 \\
& CropVLM & $78.23$ & $61.42$ & $50.28$ & $41.86$ & 46.51 & 25.79 & $50.68{\pm}0.25$ & 399.9 \\
\rowcolor{TableOurs}
& \method & $\textbf{85.71}$ & $\textbf{67.38}$ & $\textbf{54.82}$ & $\textbf{44.16}$ & \textbf{50.38} & \textbf{33.66} & $\textbf{56.02}{\pm}0.27$ & 331.9 \\
\bottomrule
\end{tabular}}
\caption{Per-benchmark results with external crop/zoom methods listed explicitly. Per-benchmark entries report mean scores over three seeds; TextVQA, DocVQA, ChartQA, and InfoVQA use 1,000-example evaluation subsets, VStarBench uses 191 examples, and MME-Lite uses 1,919 examples. Six Avg reports mean and standard deviation over the aggregate score. Bold marks the best score within each backbone and benchmark column.}
\label{tab:full-results}
\end{table*}

%% file: sections/03_method.tex
\section{Method}

\subsection{Loss-Gap Supervision}

Let $x$ denote an image, $q$ a question, and $y$ the target answer. The target VLM first receives a low-resolution global preview $g(x)$. \method learns a visual re-reading policy that predicts a preserve/review action $a \in \{\textsc{preserve}, \textsc{review}\}$, an expected utility score $u \in \mathbb{R}$, and a normalized free-form crop box $b=(c_x,c_y,w,h)\in[0,1]^4$. If the policy preserves, the model answers from the global preview. If it reviews, the system renders a local crop $c(x,b)$ and generates the final answer from the global-plus-crop context.

The supervision is mined by probing the same target VLM offline. For each training example, we construct a candidate bank $\mathcal{B}(x,q)=\{b_i\}_{i=1}^K$ with diverse centers, scales, and aspect ratios. The bank includes compact local regions and larger context-expanded regions. Each candidate is scored by comparing the target model's answer behavior under the global-only view and the crop-augmented view. For generated VQA answers, the utility of crop $b_i$ is the reduction in answer negative log-likelihood:
\[
    \Delta_i =
    \mathcal{L}_{\mathrm{NLL}}(y \mid q, g(x))
    -
    \mathcal{L}_{\mathrm{NLL}}(y \mid q, g(x), c(x,b_i)).
\]
A positive $\Delta_i$ means that the crop makes the target answer easier for the target VLM. For multiple-choice tasks, we use the correct-option margin. Let $s_y$ be the score of the correct option and $s_j$ the score of an incorrect option. The margin is
\[
    m = s_y - \max_{j\ne y} s_j ,
\]
and the crop utility is $\Delta_i=m_i-m_0$, where $m_0$ is the global-only margin and $m_i$ is the margin after injecting crop $b_i$.

The crop target and router labels come from the same utility ranking. For each example, we select
\[
    b^\star = \arg\max_{b_i \in \mathcal{B}(x,q)} \Delta_i ,
    \qquad
    \Delta^\star = \max_i \Delta_i .
\]
A clearly positive $\Delta^\star$ creates a review row: the action target is \textsc{review}, the utility target is $\Delta^\star$, and the box target is $b^\star$. Low or negative utilities create preserve rows: the action target is \textsc{preserve}, and no positive box target is applied. Ambiguous middle cases are filtered or down-weighted so that crop supervision comes from reliable answer improvements. Thus the gate, utility, and box heads are not trained from separate annotations; they are three projections of the same answer-consequence signal. For multiple-choice training, positive and negative rows are balanced so that the gate learns visual utility instead of a class prior. The labels are mined separately for each backbone because crop utility depends on the target model's visual encoder, tokenizer, decoder, prompt format, and answer parser.

\subsection{Free-Form Crop Router}

The router consumes hidden states computed from the global preview before any review crop is rendered. It predicts a gate logit for \textsc{preserve} versus \textsc{review}, a scalar utility estimate, and a continuous crop box. This pre-crop design matches deployment: the system must commit visual budget before the local crop exists, so the router learns allocation from the global state alone.

The box head uses a candidate-bank prior with bounded residual refinement. The candidate prior gives stable spatial coverage early in training, while the residual head moves the final crop continuously. The resulting box is not confined to a fixed tile grid and can follow text blocks, chart regions, document fields, or localized objects. During rendering, the predicted crop may be context-expanded when local evidence needs neighboring layout, such as a number with its unit, a chart mark with its axis, or a document value with its field label.

\subsection{Training and Inference}

Training uses dual-path examples. Preserve rows pair global-preview answering with abstention supervision, while review rows include the mined crop and supervise the gate, utility, and box. The generated answer remains in the sequence so the router is trained at the same answer boundary used at inference. The optimized router objective is
\[
\begin{aligned}
\mathcal{L} ={}&
\lambda_g \mathcal{L}_{\mathrm{gate}}
+ \lambda_u \mathcal{L}_{\mathrm{utility}}
+ \lambda_b \mathcal{L}_{\mathrm{box}} .
\end{aligned}
\]
$\mathcal{L}_{\mathrm{gate}}$ supervises the preserve/review decision, $\mathcal{L}_{\mathrm{utility}}$ regresses the mined loss-gap utility, and $\mathcal{L}_{\mathrm{box}}$ supervises the free-form crop through candidate classification, smooth continuous regression, and overlap-oriented penalties. The VLM backbone is kept fixed; only the router and its auxiliary heads are trained.

Inference is one-shot. The model encodes the global preview, the router predicts gate, utility, and box, and the system either answers immediately or injects one rendered crop. The final answer is generated in the model's standard answer format. For consistency across methods, we report scores under the same generate-and-parse protocol: normalized answer scoring for free-form VQA and option-letter scoring for multiple-choice benchmarks. We also report average visual-token usage, its ratio to the corresponding Base no-zoom setting, and the router action rate.

%% file: sections/04_experiments.tex
\section{Experiments}

\subsection{Benchmarks and Models}

We evaluate on six benchmarks that stress complementary forms of visual evidence: TextVQA~\cite{singh2019textvqa} for scene-text reading, DocVQA~\cite{mathew2021docvqa} for document reading and layout understanding, ChartQA~\cite{masry2022chartqa} for chart value extraction and relation reasoning, InfographicVQA~\cite{mathew2022infographicvqa} for text-object-layout composition, VStarBench~\cite{wu2024vstar} for fine-grained visual search and spatial reasoning, and MME-RealWorld-Lite~\cite{zhang2025mmerealworld} for high-resolution real-world multiple-choice understanding. All benchmark entries are averaged over three seeds. TextVQA, DocVQA, ChartQA, and InfographicVQA use 1,000-example evaluation subsets, VStarBench uses 191 examples, and MME-Lite uses 1,919 examples. Label mining and training use official training splits, while all reported benchmark subsets are fixed held-out validation/test examples disjoint from mining, router training, and threshold selection. The main aggregate metric is the arithmetic mean over the six benchmark scores.

We instantiate \method on three backbones: LLaVA-1.5-7B~\cite{liu2023llava}, InternVL2.5-8B~\cite{chen2024internvl}, and Qwen2-VL-2B-Instruct~\cite{wang2024qwen2vl}. TextVQA, DocVQA, ChartQA, and InfographicVQA use the free-form answer format. VStarBench and MME-Lite use the multiple-choice format for supervision and parsing.

\subsection{Label Mining and Training Data}

\input{tables/label_statistics}
\input{tables/supervision_signal_ablation}

Table~\ref{tab:label-stats} reports the source data used for label mining. Answer-likelihood labels are mined from the four free-form VQA datasets. For multiple-choice supervision, we convert GQA train-balanced questions into A--D choice rows by pairing the gold answer with answer-type-matched distractors sampled from the GQA training pool, and mine option-margin labels on these constructed choices. The retained positive rows provide review actions and crop targets; the remaining rows supply preserve supervision or are filtered according to the label rule.

Dual-path construction then creates preserve rows for global-only answering and review rows with the mined crop injected before the final answer.

\subsection{Baselines}

The main comparison includes Base no-zoom inference, external crop/zoom methods, and \method inference. \emph{Base no-zoom} uses the original VLM with one global preview. ZoomRefine, ViCrop, and CropVLM cover prompt-based zoom refinement, CLIP-guided crop selection, and learned external crop prediction under the same benchmark protocol. For VStarBench and MME-Lite, Base no-zoom and \method use the same generate-and-parse scoring.

\subsection{Main Results Across Backbones}

Table~\ref{tab:full-results} reports the full six-benchmark comparison for each backbone and method. \method improves over Base no-zoom on all reported model rows, and the task breakdown shows where the aggregate gains come from. On InternVL2.5-8B, \method raises \sixavg from 52.25 to 64.29, while the three external rows reach 54.43 (ZoomRefine), 55.84 (ViCrop), and 57.16 (CropVLM). On LLaVA-1.5-7B, \method improves over Base no-zoom by 9.81 points and exceeds ZoomRefine, ViCrop, and CropVLM. On Qwen2-VL-2B-Instruct, \method raises \sixavg from 44.65 to 56.02.

The supervision ablation keeps the backbone, router architecture, candidate bank, training schedule, and evaluation protocol fixed, changing only the teacher used to assign review labels and crop targets. This comparison isolates the source of the router signal from the benefit of adding a second visual view.

Table~\ref{tab:supervision-signal-ablation} compares loss-gap labels with two direct alternatives. Random-label supervision asks whether training a router that sometimes injects an additional crop is already enough. CLIP relevance asks whether selecting the crop most semantically related to the question is sufficient. \lossgap outperforms both controls on the four InternVL2.5-8B VQA benchmarks, improving the VQA average by 4.69 points over CLIP relevance and by 7.57 points over random labels. The result supports the central supervision choice: the useful crop is the region that improves the target answer, not merely a plausible or semantically related region.

%% file: tables/label_statistics.tex
\begin{table}[t]
\centering
\footnotesize
\setlength{\tabcolsep}{0.5pt}
\renewcommand{\arraystretch}{1.06}
\begin{tabular}{@{}>{\raggedright\arraybackslash}m{0.14\columnwidth}
                >{\raggedright\arraybackslash}m{0.36\columnwidth}
                >{\raggedleft\arraybackslash}m{0.115\columnwidth}
                >{\raggedleft\arraybackslash}m{0.115\columnwidth}
                >{\raggedleft\arraybackslash}m{0.115\columnwidth}
                >{\raggedleft\arraybackslash}m{0.115\columnwidth}@{}}
\toprule
\rowcolor{TableHeader}
Target & Source data & Rows & Pos. & Neg. & Ambig. \\
\midrule
Answer NLL & TextVQA, DocVQA,\newline ChartQA, InfoVQA & 58,311 & 23,542 & 13,679 & 21,090 \\
\addlinespace[0.15em]
Option margin & GQA-derived A--D choices & 80,000 & 21,491 & 5,496 & 53,013 \\
\bottomrule
\end{tabular}
\caption{Mined label statistics. Answer-NLL labels use TextVQA, DocVQA, ChartQA, and InfographicVQA~\cite{singh2019textvqa,mathew2021docvqa,masry2022chartqa,mathew2022infographicvqa}; option-margin labels use GQA-derived A--D choice rows~\cite{hudson2019gqa}. Positive rows provide review actions and crop targets; ambiguous rows are not box-supervised.}
\label{tab:label-stats}
\end{table}

%% file: tables/supervision_signal_ablation.tex
\begin{table}[t]
\centering
\scriptsize
\setlength{\tabcolsep}{2.6pt}
\resizebox{\columnwidth}{!}{
\begin{tabular}{lrrrrrrr}
\toprule
\rowcolor{TableHeader}
Supervision & Text & Doc & Chart & Info & Avg. & Act. & Tokens \\
\midrule
Random labels & 78.5 & 57.5 & 66.5 & 53.5 & 64.00 & 60.2\% & 410.1 \\
CLIP relevance & 83.5 & 67.5 & 67.5 & 49.0 & 66.88 & 58.0\% & 404.5 \\
\rowcolor{TableOurs}
\lossgap & \textbf{87.26} & \textbf{78.11} & \textbf{67.63} & \textbf{53.27} & \textbf{71.57} & 59.3\% & 407.6 \\
\bottomrule
\end{tabular}}
\caption{Supervision-signal ablation on the four InternVL2.5-8B VQA benchmarks. The backbone, router, candidate bank, schedule, and evaluation are fixed; only the label teacher changes. Act. is action rate and Tokens is average visual-token use.}
\label{tab:supervision-signal-ablation}
\end{table}

%% file: sections/05_analysis.tex
\section{Mechanism Analysis}
\label{sec:analysis}

\begin{figure}[t]
\centering
\includegraphics[width=\columnwidth]{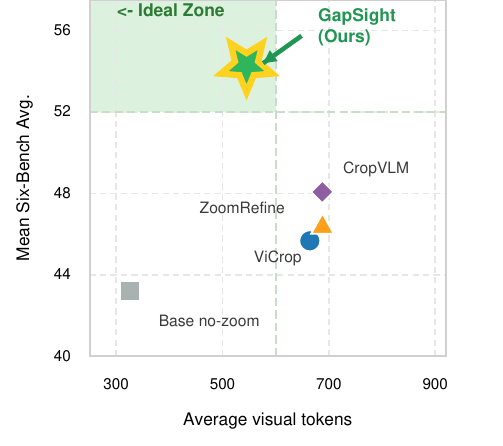}
\caption{Token-performance comparison across LLaVA-1.5-7B, InternVL2.5-8B, and Qwen2-VL-2B-Instruct. Each point averages the method's \sixavg and average visual-token usage across the three backbones. \method reaches the highest mean score while using fewer visual tokens than ZoomRefine, ViCrop, and CropVLM.}
\label{fig:token-frontier}
\end{figure}

\subsection{Useful Crops Are Model-Specific}

\method mines labels separately for each target VLM because crop utility depends on the model's own visual and language behavior. We test this assumption by transferring mined crops across backbones. For each image-question pair and candidate bank, we select the crop with the largest loss gap under a source model and then measure the realized loss gap when the same crop is evaluated by a target model. Values are normalized by the target model's own self-mined crop utility, so each within-model optimum is 1.00.

\begin{figure*}[t]
\centering
\includegraphics[width=.96\textwidth]{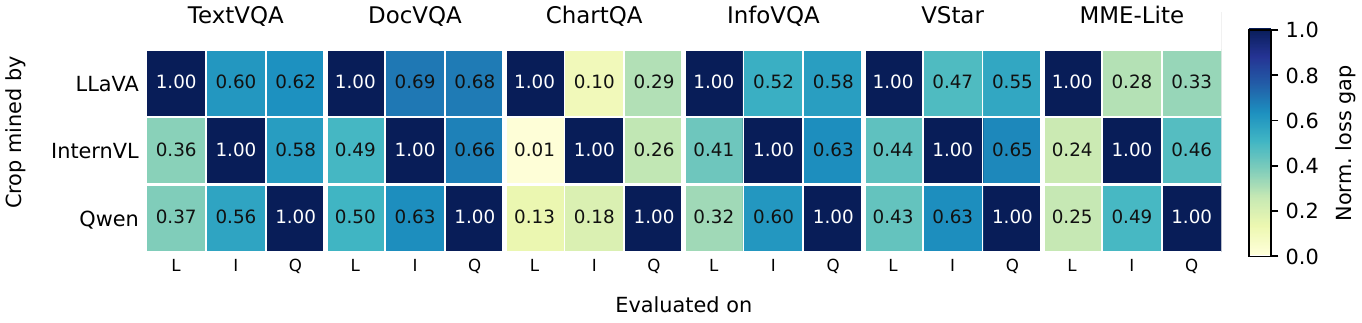}
\caption{Cross-model transfer of mined crop utility. For each benchmark, rows indicate the model used to mine the crop and columns indicate the model used to evaluate it; L, I, and Q denote LLaVA, InternVL, and Qwen. Values are normalized by the target model's self-mined crop utility. Off-diagonal values below 1.00 show that crops useful for one backbone often preserve only part of their utility for another backbone.}
\label{fig:model-specific-transfer}
\end{figure*}

Figure~\ref{fig:model-specific-transfer} shows that useful crops are only partially transferable across VLMs. Off-diagonal retention averages 0.44 across all benchmark-transfer pairs, with higher transfer on DocVQA and VStarBench but sharp drops on ChartQA and MME-Lite. The result supports model-specific label mining: the best crop is a region that changes the answer behavior of the target model, beyond a universally relevant semantic region.

This distinction matters because semantic relevance alone does not define visual utility. A crop can contain the object named in the question and still omit the neighboring text, scale cue, axis label, or scene context needed by a particular backbone. Conversely, a less obvious region can be useful if it changes the model's answer likelihood or option margin. Low off-diagonal retention therefore indicates that loss-gap mining is not simply recovering a universal question-relevant box; it captures where each backbone fails to preserve answer-critical evidence. This supports mining loss-gap labels per backbone rather than training a universal crop selector.

\subsection{When Review Helps and Hurts}

\input{tables/transition_analysis}

\method should repair Base no-zoom failures without applying crops where the global image is already sufficient. We compare Base no-zoom and \method predictions on the same InternVL2.5-8B examples from TextVQA, DocVQA, ChartQA, and InfographicVQA. Table~\ref{tab:transition} reports wrong-to-right (W$\rightarrow$R), right-to-wrong (R$\rightarrow$W), net correction, and rescue/harm rates, directly measuring whether visual re-reading creates more repairs than regressions.

\method converts 107 Base no-zoom errors into correct answers and introduces 27 regressions, giving a net correction of +80. The strongest rescue patterns appear on TextVQA and DocVQA, where local text evidence is frequently recoverable through a crop. ChartQA has fewer rescues and a lower action rate, consistent with the importance of chart-level structure. InfographicVQA shows both substantial rescues and higher harm than TextVQA, reflecting a task mixture where local text and global layout both matter.

\input{tables/gate_behavior}

The gate rates in Table~\ref{tab:gate} show that the router tracks reviewability rather than applying a fixed crop budget. It reviews frequently on text-heavy tasks, peaking at 76.5\% on InfographicVQA, but becomes much more selective when global context matters, dropping to 25.6\% on MME-Lite. This ordering matches the transition results: tasks dominated by local evidence receive frequent review, while chart and global reasoning tasks use crops more sparingly. On MME-Lite, this selective operating point raises the score from 29.66 under Base no-zoom to 40.76.

The conditional behavior explains the fragility of always-crop policies. Reviewing every example helps many OCR cases but removes useful context on globally grounded questions. Reviewing rarely preserves cost but misses small local evidence. The learned router occupies the useful middle regime: it reviews frequently where detail is decisive and abstains where the global view carries essential context.

Figure~\ref{fig:token-frontier} summarizes the score--cost frontier after averaging over the three reported backbones. \method occupies the favorable operating point: it improves over Base no-zoom and exceeds the external crop/zoom baselines while using fewer visual tokens. The frontier separates selective routing from brute-force visual-token expansion. External crop and zoom baselines spend additional visual budget broadly, whereas \method spends that budget only when the global state predicts answer-consequential local evidence. The gain therefore comes from budget allocation, not from giving every example a larger visual input.

This allocation view also connects the gate, utility, and box predictions. The gate controls whether extra tokens are spent, the utility estimates answer benefit, and the box determines which local evidence enters the second view. Because preview token budgets differ across backbones, action rate alone is not a faithful cost measure. The token frontier indicates that \method learns this coupling: it spends crop tokens on high-utility cases while avoiding the broad second-pass cost paid by always-crop or prompt-driven zooming methods.

\subsection{Crop Geometry and Context}

The mined and predicted boxes are continuous. In the InternVL label audits, VQA positives have median area about 0.20, and VStar/MME positives have median area about 0.20, with upper quantiles around 0.55--0.56. The reported InternVL operating point has mean predicted area around 0.315. These areas preserve context around local evidence while reducing the visual search space compared with repeating the full image.

This geometry matters for text-rich tasks. A number needs its unit, a chart mark needs its axis, and a document value needs its field label. \method predicts evidence boxes and renders context-expanded crops when needed, giving the model a local view that remains connected to the surrounding layout. The box head therefore learns answer-consequential regions rather than generic saliency: it spends crop tokens where local detail can change the model's prediction while retaining enough surrounding context to make that detail usable. The resulting crop is an answer context rather than a detector box, small enough to recover detail but broad enough to preserve the relation that makes the detail interpretable.

%% file: tables/transition_analysis.tex
\begin{table}[t]
\centering
\small
\setlength{\tabcolsep}{4pt}
\begin{tabular}{lrrrrr}
\toprule
\rowcolor{TableHeader}
Benchmark & W$\rightarrow$R & R$\rightarrow$W & Net & Rescue & Harm \\
\midrule
TextVQA & 28 & 5 & +23 & 59.6 & 3.3 \\
DocVQA & 44 & 7 & +37 & 55.7 & 5.8 \\
ChartQA & 9 & 6 & +3 & 13.6 & 4.5 \\
InfographicVQA & 26 & 9 & +17 & 23.9 & 9.9 \\
\midrule
All & 107 & 27 & +80 & 35.6 & 5.4 \\
\bottomrule
\end{tabular}
\caption{Benefit/harm transition analysis on the four InternVL2.5-8B VQA benchmarks. W$\rightarrow$R counts Base no-zoom errors corrected by \method; R$\rightarrow$W counts regressions. Rescue and harm are percentages normalized by Base no-zoom wrong/right examples.}
\label{tab:transition}
\end{table}

%% file: tables/gate_behavior.tex
\begin{table}[t]
\centering
\small
\setlength{\tabcolsep}{4pt}
\begin{tabular}{lrrr}
\toprule
\rowcolor{TableHeader}
Benchmark & Score & Action & Avg. tokens \\
\midrule
TextVQA & 87.26 & 0.740 & 445.4 \\
DocVQA & 78.11 & 0.515 & 387.8 \\
ChartQA & 67.63 & 0.350 & 345.6 \\
InfographicVQA & 53.27 & 0.765 & 451.8 \\
VStarBench & 58.72 & 0.539 & 394.1 \\
MME-Lite & 40.76 & 0.256 & 321.5 \\
\bottomrule
\end{tabular}
\caption{Task-level gate behavior for the InternVL2.5-8B reported operating point. The router reviews frequently on OCR/infographic tasks and is more conservative on MME-Lite.}
\label{tab:gate}
\end{table}

%% file: sections/07_conclusion.tex
\section{Conclusion}

We introduced \method, a framework for learning visual re-reading from model-specific loss gaps. By probing candidate crops offline and measuring answer-loss or option-margin improvements, \method converts a VLM's own answer behavior into supervision for when and where to look again. A lightweight free-form crop router distills these signals into a one-shot action, utility score, and continuous crop box. Across multiple VLM backbones and six benchmarks, \method improves Base no-zoom inference and reaches strong aggregate performance against recent crop and zoom baselines. The mechanism analyses show that loss-gap labels yield task-adaptive and model-specific visual behavior: the router rescues concrete errors, adjusts its action rate by task, predicts context-preserving boxes, and improves the token-performance profile.